\documentclass[sigconf]{acmart}
\renewcommand\footnotetextcopyrightpermission[1]{} 

\usepackage{booktabs} 

\usepackage{url,bm}
\usepackage{latexsym}

\usepackage{algorithm}
\usepackage[noend]{algorithmic}
\usepackage{multirow}

\usepackage{etoolbox}
\newtoggle{conf} 
\toggletrue{conf} 

\newcommand{\cut}[1]{}

\copyrightyear{2018} 
\acmYear{2018} 
\setcopyright{rightsretained} 
\acmConference[SIGIR '18]{The 41st International ACM SIGIR Conference on Research \& Development in Information Retrieval}{July 8--12, 2018}{Ann Arbor, MI, USA}
\acmBooktitle{SIGIR '18: The 41st International ACM SIGIR Conference on Research \& Development in Information Retrieval, July 8--12, 2018, Ann Arbor, MI, USA}
\acmDOI{10.1145/3209978.3210205}
\acmISBN{978-1-4503-5657-2/18/07}

\begin{document}
\title{Talent Search and Recommendation Systems at LinkedIn:\\ Practical Challenges and Lessons Learned}

\author{Sahin Cem Geyik, \ Qi Guo, \ Bo Hu, \ Cagri Ozcaglar, \ Ketan Thakkar, \\ Xianren Wu, \ Krishnaram Kenthapadi}
\orcid{1234-5678-9012}
\affiliation{%
  \institution{LinkedIn Corporation, USA}
}



%
%

\keywords{Talent Search \& Recommendation, Candidate Retrieval \& Ranking}

\maketitle

\section{\large{Talent Search and Recommendation: Practical Challenges}}\label{sec:talentsearch}

\let\thefootnote\relax\footnote{\\ ~ \\ ~  \large \textbf{This paper has been accepted for publication at ACM SIGIR 2018.}}LinkedIn Talent Solutions business contributes to around 65\% of LinkedIn's annual revenue, and provides tools for job providers to reach out to potential candidates and for job seekers to find suitable career opportunities. LinkedIn's job ecosystem has been designed as a platform to connect job providers and job seekers, and to serve as a marketplace for efficient matching between potential candidates and job openings. A key mechanism to help achieve these goals is the \emph{LinkedIn Recruiter} product, which enables recruiters to search for relevant candidates and obtain candidate recommendations for their job postings.

We highlight a few unique information retrieval, system, and modeling challenges associated with talent search and recommendation systems:
\begin{enumerate}
\item The underlying query to the talent search system could be quite complex, combining several structured fields (such as canonical title(s), canonical skill(s), company name) and unstructured fields (such as free-text keywords). Depending on the application, the query could either consist of an explicitly entered query text and selected facets (talent search), or be implicit in the form of a job opening, or ideal candidate(s) for a job (talent recommendations). Our goal is to determine a ranked list of most relevant candidates in real-time among hundreds of millions of structured candidate profiles. Consequently, robust standardization, efficient indexing, candidate selection, and multi-pass scoring/ranking systems are essential \cite{galene_engine, thucS16ltr}.
\item Unlike traditional search and recommendation systems which solely focus on estimating how relevant an item is for a given query, the talent search domain requires mutual interest between the recruiter and the candidate in the context of the job opportunity. In other words, we require not just that a candidate shown must be relevant to the recruiter's query, but also that the candidate contacted by the recruiter must show interest in the job opportunity. Hence, it is crucial to use appropriate metrics (e.g., the likelihood of a candidate receiving an inMail (message) from the recruiter and also answering with a positive response) for model optimization as well as for online A/B testing, taking into account the fact that certain ideal metrics (e.g., the likelihood of a candidate receiving a job offer and accepting it) may either be unavailable or delayed \cite{Ramanath_2018, thuc15pes}.
\item Quite often, the recruiter or the hiring manager may not be able to express their hiring needs in the form of a search query (or even a job posting), since this often requires deep domain knowledge, as well as significant time and manual effort to come up with the best search criteria (e.g., which skills are relevant for a specific role that the recruiter is looking to fill). To address this challenge, it is desirable to support search based on ideal candidate(s) \cite{Thuc16}, and online learning of recruiter preferences within a search session based on their instantaneous response to recommended candidates \cite{Geyik_2018}.
\end{enumerate}
In this talk, we will present how we formulated and addressed the above problems, the overall system design and architecture, the challenges encountered in practice, and the lessons learned from the production deployment of these systems at LinkedIn. By presenting our experiences of applying techniques at the intersection of recommender systems, information retrieval, machine learning, and statistical modeling in a large-scale industrial setting and highlighting the open problems, we hope to stimulate further research and collaborations within the SIGIR community.

\section{\large{Overview of Talent Search and Recommendation Systems Developed and Deployed at LinkedIn}}

\begin{figure} [!h]
\centering
\includegraphics[width=3.3in]{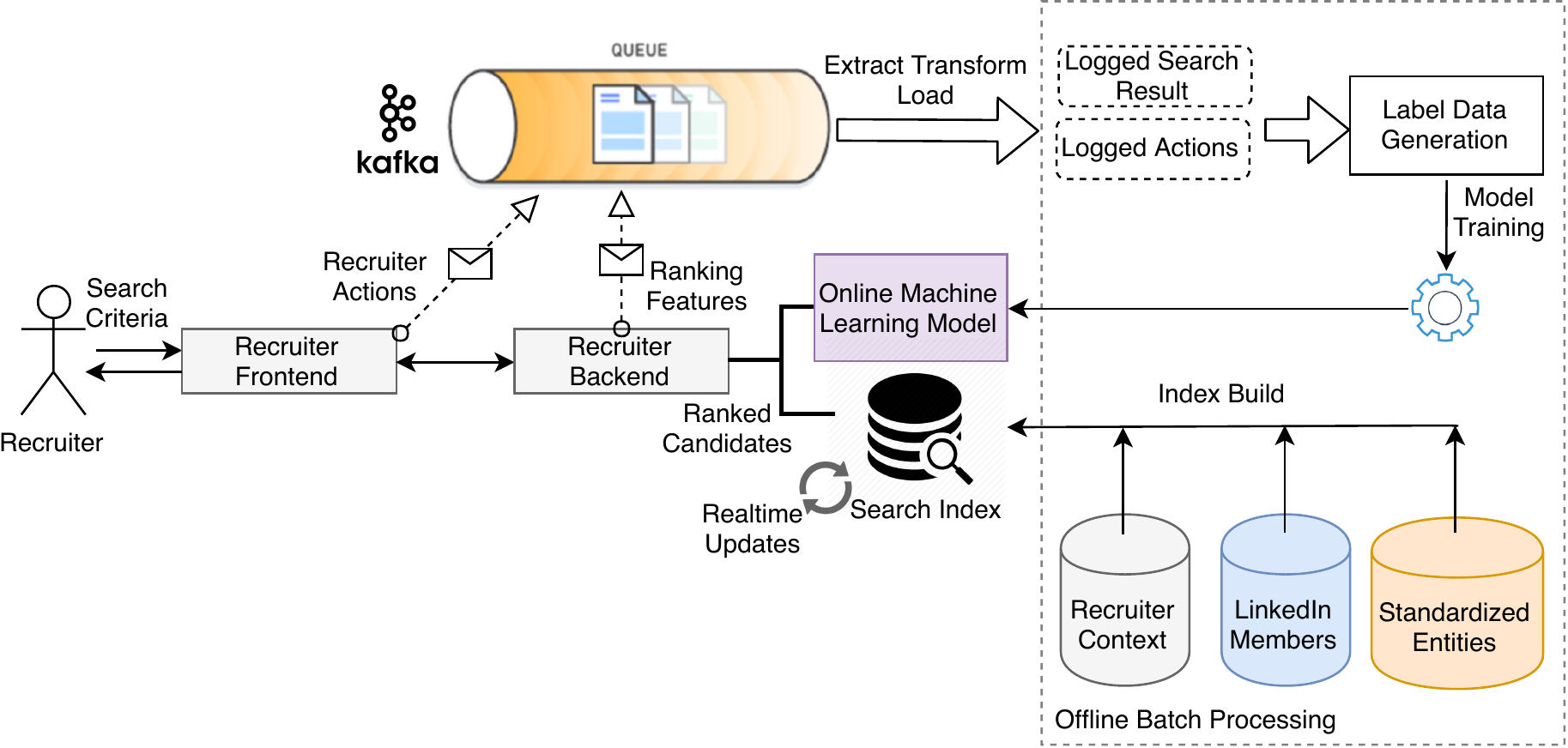} 
\caption{Architecture of LinkedIn's talent search and recommendation engine.}
\label{fig:RecruiterSearchInfra}
\end{figure}

\begin{figure} [!h]
	\centering
	\includegraphics[width=3.3in]{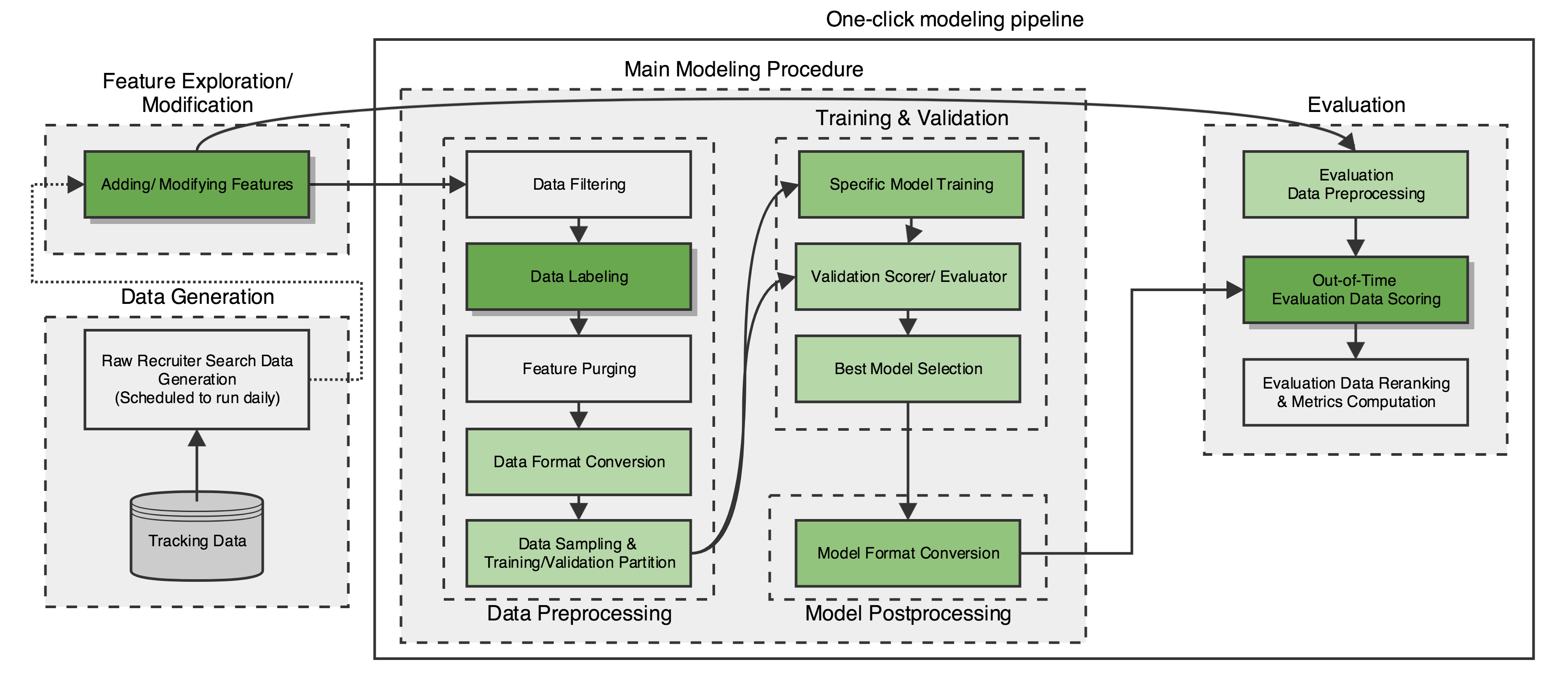} 
	\caption{Offline modeling pipeline for LinkedIn's talent search and recommendation engine.}
	\label{fig:recruiter-search-offline-pipeline}
\end{figure}

We next briefly describe the overall architecture of LinkedIn's talent search and recommendation engine, highlighting the key components (Figure~\ref{fig:RecruiterSearchInfra}). Our system can be subdivided into an online system for serving most relevant candidate results and an offline workflow for updating different machine learned models (described in greater detail in Figure~\ref{fig:recruiter-search-offline-pipeline}). Our presentation covers the architecture choices, modeling design decisions, and the practical lessons learned.

{\em Online system architecture}: First, the recruiter's search request (either as explicitly entered query, or implicit in the form of job opening / ideal candidate(s)), along with the recruiter and session context, is transformed into a complex query combining structured fields (e.g., canonical title(s) / skill(s), company name, region) and unstructured text keywords, and issued to LinkedIn's {\em Galene} search engine \cite{galene_engine}. A candidate set of results is then retrieved from the search index based on the criteria specified, and then ranked in multiple passes using machine learned scoring models of varying complexity \cite{thucS16ltr, Ramanath_2018, thuc15pes, Thuc16, Geyik_2018}. The search result set, along with the features used by the ranking model, are logged for later use for model training. Finally, front-end server gets the top ranked candidates, renders the result page, and logs recruiter interactions. The underlying search index is updated in near real-time to reflect changes in LinkedIn member data.

{\em Offline modeling pipeline}: Our offline system periodically trains the ranking models using recruiter usage logs \cite{thucS16ltr, Ramanath_2018, thuc15pes, Thuc16, Geyik_2018}. The training data is generated from recruiter interactions (and candidate responses to recruiter messages) over the search results displayed. As the member data can change over time, we also log computed features along with search results, instead of generating the features during model training. The offline modeling pipeline is designed to support ease of feature engineering, incorporation of different types of machine learning models, and experimentation agility.

\section{\large{Author Bios}}\label{sec:bio}
The authors have extensive experience in the field of web-scale search and recommendation systems, and in particular, in applying data mining, machine learning, and information retrieval techniques in the talent search domain. They have built and deployed multiple generations of machine learning models and systems for real-time, low latency applications such as talent search and recommendations at LinkedIn. They have published extensively in venues such as SIGIR, KDD, WWW, WSDM, and CIKM, and also presented tutorial/industry talks about their work.

Sahin Cem Geyik is part of the AI team at LinkedIn, focusing on personalized recommendations across several LinkedIn Talent Solutions products. He received his Ph.D. degree in Computer Science from Rensselaer Polytechnic Institute in 2012, and has authored papers in top-tier conferences and journals such as KDD, INFOCOM, IEEE TMC, and IEEE TSC.

Qi Guo is part of the AI team at LinkedIn, where he applies machine learning for LinkedIn Talent Solutions products. He received his M.S. degree in Robotics from Carnegie Mellon University in 2016. He has published at IJCAI.

Bo Hu is part of the AI team at LinkedIn, where he works on relevance for LinkedIn Talent Solutions products. He received his Ph.D. degree in Computer Science from Simon Fraser University in 2014, and has authored papers in top-tier conferences and journals such as RecSys, ICDM, IEEE TKDE, and ACM TOIS.

Cagri Ozcaglar is part of the AI team at LinkedIn, where he works on relevance for LinkedIn Talent Solutions products. He received his Ph.D. degree in Computer Science from Rensselaer Polytechnic Institute in 2012, and has authored papers in top-tier conferences and journals such as IEEE BIBM, IEEE TNBS, BMC Genomics, Mathematical Biosciences.

Ketan Thakkar is a relevance engineer at LinkedIn Talent Solutions and works on improving search relevance on LinkedIn search stack. Previously, he was part of Microsoft's Bing search relevance team working on improving relevance for Bing search and ads. He received his M.S. degree in Information Technology from Bentley University in 2010.

Xianren Wu is part of the AI team at LinkedIn, leading the candidate recommendation relevance efforts within LinkedIn Talent Solutions. He previously co-founded and was the director of R\&D for GageIn Inc. He received his Ph.D. degree in Electrical Engineering from U.C. Santa Cruz in 2008, and has authored papers in top-tier conferences such as WWW and CIKM.

Krishnaram Kenthapadi is part of the AI team at LinkedIn, where he leads the fairness and privacy modeling efforts across different LinkedIn applications. Previously, he was a Researcher at Microsoft Research Silicon Valley. He received his Ph.D. degree in Computer Science from Stanford University in 2006. He has published 35+ papers, filed 125+ patents, and received the CIKM best case studies paper award, the SODA best student paper award, and the WWW best paper award nomination.

{\em About LinkedIn}: Founded in 2003, LinkedIn connects the world's professionals to make them more productive and successful. LinkedIn operates the world's largest professional network on the Internet with more than 500 million members in over 200 countries and territories. The company has a diversified business model with revenues coming from talent solutions, marketing solutions, and premium subscription products. See \url{https://press.linkedin.com/about} for more information.

\bibliographystyle{ACM-Reference-Format}
\bibliography{paper}

\end{document}